\documentclass[10pt,twocolumn,letterpaper]{article}

\usepackage[pagenumbers]{cvpr} 

\usepackage{graphicx}
\usepackage{amsmath}
\usepackage{amssymb}
\usepackage{booktabs}

\usepackage[pagebackref,breaklinks,colorlinks]{hyperref}

\usepackage{enumitem}
\usepackage{array}
\usepackage{balance}
\usepackage{color}
\usepackage{bbm}
\usepackage{algorithm}
\usepackage[noend]{algpseudocode} 
\usepackage{bbding}
\usepackage{booktabs}
\usepackage[normalem]{ulem}
\usepackage{setspace}
\usepackage{stfloats}
\usepackage{bm}

\usepackage{amsmath}
\usepackage{amssymb}
\usepackage{amsfonts}       
\usepackage{cite}

\newcolumntype{C}[1]{>{\centering\let\newline\\\arraybackslash\hspace{0pt}}m{#1}}

\def\cvprPaperID{23} 
\def\confName{CVPR}
\def\confYear{2022}

\begin{document}

\title{AuxMix: Semi-Supervised Learning with Unconstrained Unlabeled Data}


\author{Amin Banitalebi-Dehkordi, Pratik Gujjar, and Yong Zhang\\
Huawei Technologies Canada Co., Ltd.\\
Vancouver, BC, Canada\\
{\tt\small \{amin.banitalebi, pratik.gujjar, yong.zhang3\}@huawei.com}
}

\maketitle

\begin{spacing}{1.00}

\begin{abstract}

Semi-supervised learning (SSL) has seen great strides when labeled data is scarce but unlabeled data is abundant. Critically, most recent work assume that such unlabeled data is drawn from the same distribution as the labeled data. In this work, we show that state-of-the-art SSL algorithms suffer a degradation in performance in the presence of unlabeled \textit{auxiliary} data that does not necessarily possess the same class distribution as the labeled set. We term this problem as \textit{Auxiliary-SSL} and propose \textit{AuxMix}, an algorithm that leverages self-supervised learning tasks to learn generic features in order to mask auxiliary data that are not semantically similar to the labeled set. We also propose to regularize learning by maximizing the predicted entropy for dissimilar auxiliary samples. We show an improvement of 5\% over existing baselines on a ResNet-50 model when trained on CIFAR10 dataset with 4k labeled samples and all unlabeled data is drawn from the Tiny-Imagenet dataset. We report competitive results on several datasets and conduct ablation studies. 

\end{abstract}

\section{Introduction}
\label{sec:intro}






Collecting and annotating large amounts of labeled data remains to be a fundamental barrier in exploiting the full potential of machine learning algorithms. Not only is it time consuming and costly, in some cases (e.g. rare medical conditions) it might not even be possible. To mitigate these issues, there are a number of semi-supervised learning (SSL) algorithms proposed in the literature which aim to supplement a small set of available labeled data with a larger set of unlabeled examples during training \cite{van2020survey,s4l,fixmatch-2020,mixmatch,banitalebi2021revisiting}. Unlabeled data, when properly used in conjunction with the available labeled data, can greatly improve the learning generalization \cite{selflabelsurvey,banitalebi2021knowledge}.

The majority of the existing semi-supervised learning approaches assume that the unlabeled data is drawn from the same distribution as the labeled training dataset. In many cases where the unlabeled data is cheaply available, this does not impose any problems. In fact, many state-of-the-art (SOTA) SSL algorithms can be used to reduce the need for manual labeling of the data. However, this is a strong assumption, and will not necessarily always hold. In line with the observation made by Oliver et al., \cite{oliver}, we will show that existing SSL algorithms suffer a considerable performance drop if the unlabeled dataset is chosen from sources other than the same distribution as the labeled set. For example, FixMatch- with 95.7\% top-1 classification accuracy when using 4K CIFAR10 labeled and 46K CIFAR10 unlabeled examples \cite{fixmatch-2020}, drops down to only 58.48\% accuracy when using 4K CIFAR10 labels but 100K Tiny-Imagenet \cite{le2015tiny} unlabeled examples (more evidence and results in Section \ref{sec:experiments}). The significant drop is due to the unlabeled distribution mismatch in the two cases. Intuitively this is also expected as any information (such as pseudo-labels) about the unlabeled data queried from a model trained with a different training label distribution will not be reliable. In addition, in some applications where collecting unlabeled data in large portions may not be possible, such data are available for related applications (e.g. neck x-ray vs knee x-ray). This necessitates new solutions to address the aforementioned problems.



\begin{figure*}
\centering
    \vspace{8pt}
    \includegraphics[width=1.6\columnwidth]{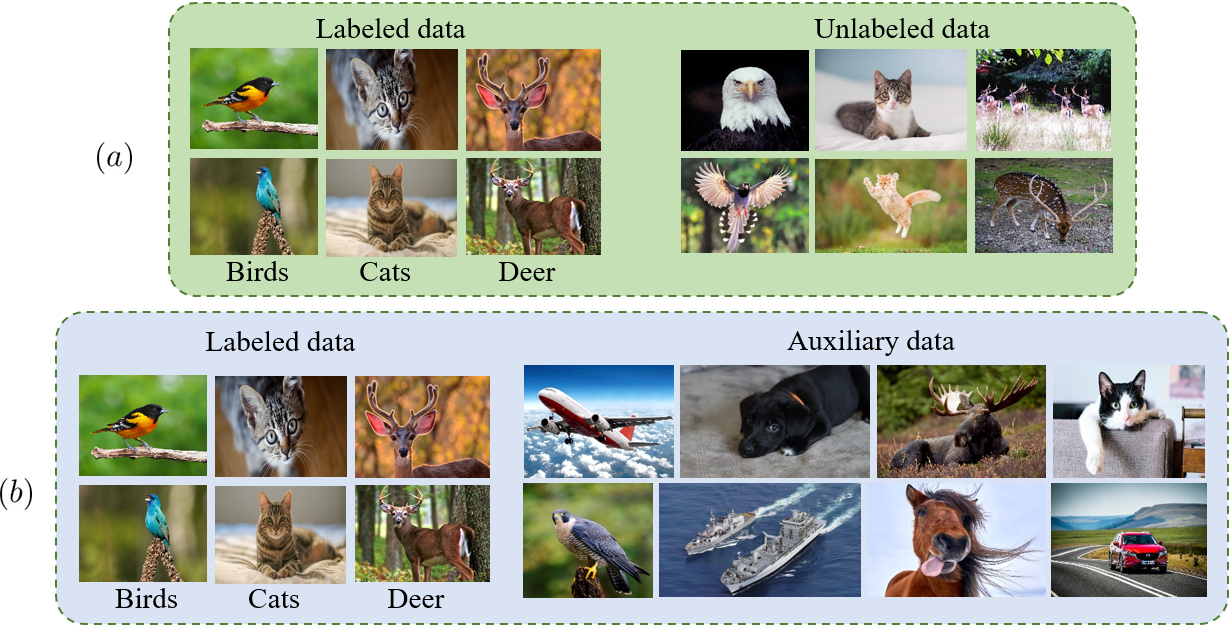}
    \caption{A visualization of unlabeled versus auxiliary data in the semi-supervised setting. (a) Unlabeled data is from the same distribution as the labeled data. (b) Unlabeled data contains out-of-distribution data.}%
    \vspace{8pt}
    \label{fig:problem-visualization}%
\end{figure*}

In this paper, we formalize this problem and introduce it as a new paradigm termed Auxiliary-SSL. We refer to the unlabeled data from unconstrained distributions as auxiliary data to distinguish it with the in-distribution unlabeled data settings used in existing SSL approaches (See Figure \ref{fig:problem-visualization} for a visualization). 
In addition, we propose AuxMix, an algorithm of mixing the labeled and auxiliary data in an efficient way that reduces the degradation gap caused in Auxiliary-SSL.

The main contributions of this paper can be summarized as follows:

\begin{itemize}
    \item We argue that the existing semi-supervised learning algorithms will under-perform, sometimes at big margins, if they use unlabeled data from unconstrained distributions (auxiliary data), and confirm this empirically on multiple datasets.
    \item We establish this paradigm as Auxiliary-SSL, and provide a formal definition for it.
    \item We propose AuxMix, a method for tackling the Auxiliary-SSL. Our method makes use of self-supervised tasks to learn an initial representation of the auxiliary data, with which we then mask the auxiliary set. We also propose a technique of entropy maximization for the dissimilar negative auxiliary samples. Experimentally, we confirm that our method is effective in reducing the performance drop caused by the label distribution mismatch of the Auxiliary-SSL.
\end{itemize}

The rest of the paper is organized as follows. Section \ref{sec:related_work} provides a brief summary of the related literature. Section \ref{sec:problem_setting} formalizes the Auxiliary-SSL paradigm. Section \ref{sec:method} describes our AuxMix method in details. Section \ref{sec:experiments} contains the experiment results and discussions around the observations. Finally, Section \ref{sec:conclusion} concludes the paper.

\section{Related Work}
\label{sec:related_work}


In this section, we provide a brief summary of literature related to our work. Broadly, we look at state-of-the-art semi-supervised learning approaches, some recent works that investigate label distribution mismatch and open-set domain adaptation methods.

\subsection{Semi-supervised learning}
There is an extensive variety of literature on semi-supervised learning algorithms, especially after the boom of deep learning \cite{selflabelsurvey}. Among them, pseudo-label based methods \cite{sohn2020simple,arazo2020pseudo,rizve2021defense,lokhande2020generating,banitalebi2021repaint,ramamonjison2021simrod} train a model on the existing labeled data and then use this model to generate pseudo-labels on the unlabeled data, which will later be used for additional training. Another emerging direction is to leverage self-supervised learning algorithms such as RotNet \cite{rotnet-2018}, JigSaw \cite{jigsaw}, SimCLR \cite{simclr}, or MOCO \cite{moco} for unsupervised pretraining and then fine-tune with the limited labeled set \cite{kim2021selfmatch, simclr-v2}. Furthermore, a number of recent SOTA methods rely on regularizing consistency in predictions across different augmentations \cite{bachman2014learning}. $\Pi$-model retains an exponential-moving-average (EMA) of the predictions \cite{TemporalEnsembling} whereas Mean Teacher proposes to average model parameters instead \cite{meanteacher-2017, tarvainen2017mean}. In an other line of work, Virtual Adversarial Training (VAT) defines adversarial perturbations as input augmentations \cite{miyato2018virtual}. \cite{verma2019interpolation, mixup} in their work enforce a consistency between convex interpolations of unlabeled samples and their similarly interpolated predictions. 
Using MixUp \cite{mixup} as their augmentation procedure, MixMatch \cite{mixmatch} and ReMixMatch \cite{remixmatch} also perform consistency regularization. 
Our work is also closely related to UDA \cite{uda} and FixMatch \cite{fixmatch-2020}. Both of these methods use predictions from a weakly augmented sample to regularize prediction on a strongly augmented sample. ReMixMatch \cite{remixmatch} also follows a similar procedure along with Distribution Alignment and Augmentation Anchoring. UDA further sharpens the predicted distribution whereas FixMatch uses pseudo-labels instead. These methods rely on confidence-based thresholding to filter spurious predictions. As mentioned in Section \ref{sec:intro}, in the presence of unconstrained auxiliary data, the traditional semi-supervised approaches suffer a considerable performance drop due to improper filtering of auxiliary data coupled with unreliable predictions for the out-of-distribution unlabeled set.

\subsection{Class distribution mismatch}
As discussed in section \ref{sec:intro}, the problem of class distribution mismatch between labeled and unlabeled data is more practically observed. Efforts in this area however, are limited. Oliver et al., \cite{oliver} in their paper draw forth several scenarios for evaluating semi-supervised learning.  They show that adding unlabeled data from a mismatched set of classes can actually degrade the performance compared to not using them at all. RealMix \cite{realmix} tackles this problem with consistency training and Entropy Minimization. They show their results using 400 labeled samples per class from CIFAR-10 Animals subset with the rest of CIFAR-10 classes as the mismatched unlabeled set. Chen et al., \cite{uasd} propose Uncertainty-Aware Self-Distillation (UASD) that uses soft-targets and self-distillation to combat class mismatch. DS3L \cite{ds3l} provide a theoretical guarantee that their algorithm never learns worse than with only using labeled data. They showcase their results on MNIST and CIFAR10 datasets with upto 60\% mismatch. In this paper, we formalize this problem as Auxiliary-SSL and refer to the unconstrained unlabeled data as auxiliary set. We report improved results when using Tiny-Imagenet, Caltech-UCSD Birds (CUB) \cite{cub} and noise images drawn from a uniformly random distribution as auxiliary datasets.

\begin{figure*}[!b]
\centering
    \vspace{10pt}
    \includegraphics[width=1.3\columnwidth]{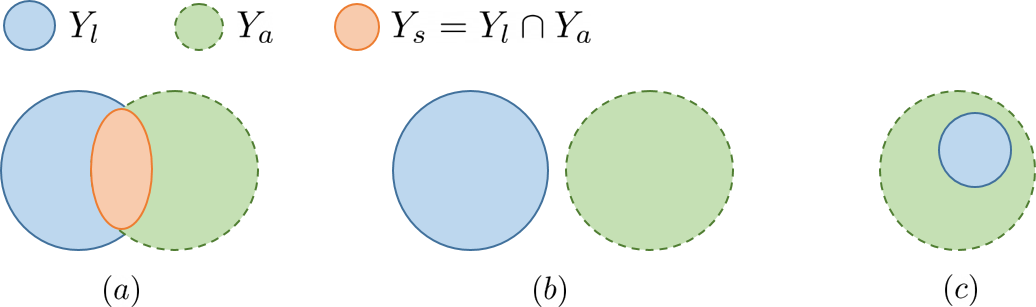}
    \caption{Venn diagram illustrating the relationship between two labeled and auxiliary distributions: (a) partial overlap, (b) no overlap, and (c) complete overlap.}%
    \label{fig:lab-aux-venn}%
\end{figure*}

\subsection{Open-set domain adaptation techniques}
Domain adaptation (DA) methods also have a similar setup to Auxiliary-SSL. The goal of DA is to make accurate predictions on the target domain having learnt from labeled data in the source domain. In contrast, Auxiliary-SSL tries to enhance learning from limited labeled data in the source domain itself by leveraging related data from an unlabeled dataset. Also, traditional (closed-set) domain adaptation deals with the problem of a distribution shift between the source and the target domain but assumes that the labels remain fixed \cite{ben2007analysis}, which is different than our Auxiliary-SSL setting. That being said, open-set domain adaptation does consider the case of label distribution mismatch between the source and target domain datasets \cite{panareda2017open,saito2018open,liu2019separate}.  However unlike the Auxiliary-SSL unconstrained paradigm, these methods still impose some conditions on the unlabeled dataset used for adaptation. For example, they usually assume a set of known target classes along with a set of unknown categories. In addition, the evaluation of such methods is usually in controlled scenarios where some classes of a dataset are considered only for the target dataset. On the contrary, in our method, we relax these constraints (even experiment with training with pure noise in Section \ref{sec:experiments}), and allow the auxiliary set to be freely selected. To cope with the potential problems of this design choice, we propose a method to filter the auxiliary examples and regularize the negatively masked samples.

\section{Problem Setting}
\label{sec:problem_setting}

We formulate our problem of learning from auxiliary unlabeled data following that of existing semi-supervised learning (SSL) work \cite{oliver}. A key change in our work is the more general assumption that the data from the unlabeled set does not share the same class distribution as the labeled set. Therefore, a classifier trained on the labeled set cannot trivially generate labels for the unlabeled set. We denote this paradigm as Auxiliary-SSL, and propose AuxMix as a method to tackle this problem. We refer to the unlabeled set as the auxiliary set with samples drawn from an auxiliary distribution.

More formally, we are given a labeled set $D_{l} = \{(x_{i}^{l}, y_{i}^{l}) \sim p\}_{i=1}^{n_{l}}$ sampled i.i.d. from an unknown data generating distribution $p(X, Y_{l})$ and an auxiliary set $D_{a} = \{(x_{i}^{a}) \sim q_{x}\}_{i=1}^{n_{a}}$ sampled i.i.d from distribution $q(X)$ which is a marginalization of the data distribution over labels in the auxiliary domain $Y_{a}$. The shared label set between the two distributions $Y_{s} = Y_{l} \cap Y_{a} \neq \varnothing$ and the auxiliary set has private labels $Y_{p} = Y_{a} \backslash Y_{s}$.

It is worth noting that in general, the relationship between two label distributions $Y_{l}$ and $Y_{a}$ can be characterized in the following different ways as shown in Figure \ref{fig:lab-aux-venn}: (a) The label set $Y_{l}$ could have a partial overlap with $Y_{a}$, (b) no overlap between $Y_{a}$ and $Y_{l}$, and (c) a complete overlap between $Y_{l}$ and $Y_{a}$ reverting to standard semi-supervised learning setting. In this work, we consider the first two cases. The objective of our algorithm is therefore to learn a model $f_{\theta}$ that maximizes the classification performance on the source labeled set $Y_{l}$ by distinguishing between auxiliary samples in the shared label set $Y_{s}$ and the auxiliary private label set $Y_{p}$.

\section{Method}
\label{sec:method}
Our algorithm, dubbed AuxMix, consists of two phases. In phase one, we learn to score each auxiliary data sample $\{(x_{i}^{a}) \sim D_{a}\}^{n_{a}}_{i=1}$ according to their semantic similarity with labeled data and separate them into a positive selection set $D^{+}_{a}$ and a negative regularization set $D^{-}_{a}$. In phase two, the positive set $D^{+}_{a}$ is used in semi-supervised learning based on consistency regularization across augmentations \cite{fixmatch-2020}. We further regularize learning from auxiliary data by maximizing the entropy of predictions on the negative set $D^{-}_{a}$ to penalize confident output distributions \cite{entmax-hinton-2017}. Intuitively, the positive set encourages reliable predictions for a meaningful consistency regularization. On the other hand, the model is required to be unable to decide which class labels to assign the samples in the negative set. We do this by pushing the model towards high uncertainty for these data samples.

\subsection{Scoring and pre-training based on self-supervised learning}
\label{sec:phase1}
We train a model $f_{\theta}^{p}$ consisting of an encoder $g^{p}(.)$ and a classifier head $c^{p}(.)$. In the limited labeled data setting of Auxiliary-SSL, we employ self-supervised learning to pre-train the model on a self-supervised prediction task. Self-supervised tasks allow the model to learn general representations from the data \cite{jing2020self}. In the absence of ground-truth labels for the auxiliary samples, these representations can be exploited to gain useful insights about the samples. There are several self-supervised learning tasks in recent literature that have shown promising results in the complete absence of labels as discussed in section \ref{sec:related_work}. In AuxMix, we leverage rotation prediction following the work of Gidaris et al., \cite{rotnet-2018} that shows that by predicting the rotations for each image following their augmentation, the model learns to extract semantic features of objects contained in them. These semantic features become key to measuring similarity between the images of the auxiliary set and the labeled set at a later stage. However, other self-supervised approaches could also be utilized.

\begin{figure*}
\centering
    \vspace{16pt}
    \includegraphics[width=1.8\columnwidth]{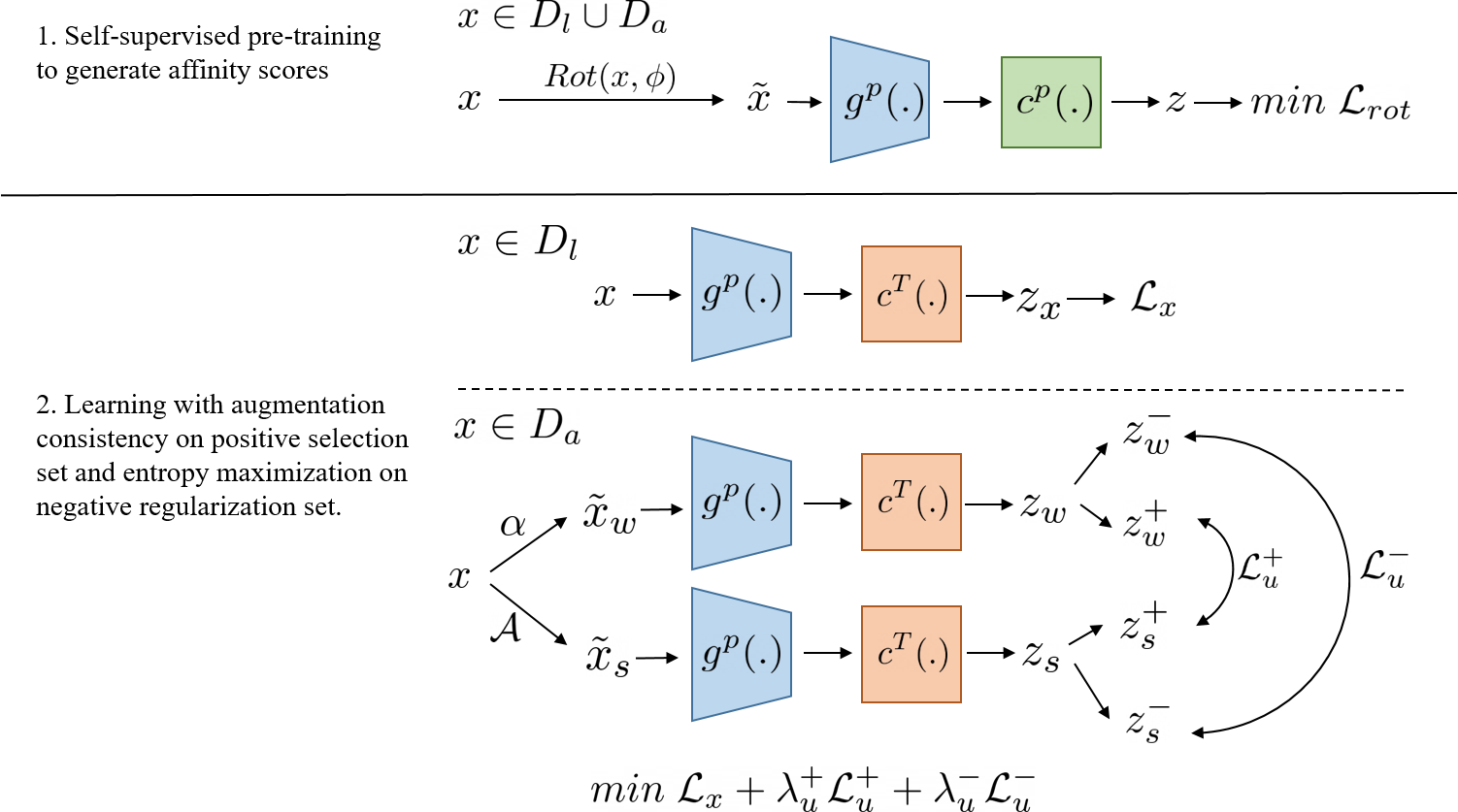}
    \caption{Visual description of the proposed AuxMix method. In Phase 1, we pre-train a model on self-supervised rotation and use the representations to generate affinity scores for the auxiliary dataset. In phase 2, we perform consistency regularization on the positive samples and entropy maximization on the negative samples at the same time as learning from the limited labeled data. The overall loss will therefore have three terms.}%
    \label{fig:algo-pic}%
    \vspace{8pt}
\end{figure*}

Given a set of training images $\{x_{i}^{p} \sim D_{l} \cup D_{a}\}^{n_{l}+n_{a}}_{i=1}$ the model $f_{\theta}^{p}$ is first trained to minimize the rotation prediction loss $\mathcal{L}_{rot}$ defined as:
\begin{equation}
\begin{split}
    \quad \quad \mathcal{L}_{rot} = H\big(\phi, f_{\theta}(Rot(x_{i}^{p}, \phi)\big) & \\ \forall \; \phi \in [0^{\circ}, \; 90^{\circ}, & \; 180^{\circ}, \; 270^{\circ}],
\end{split}
\end{equation}
where, $Rot(x_{i}, \phi)$ is an augmentation operator that rotates the images by the specified angle $\phi$ and $H(y, f(x))$ is the standard cross-entropy loss between a target $y$ and a prediction $f(x)$. We then calculate a benchmark semantic representation or a prototype for each class $c$ in $Y_{l} = [1, \dots, C]$ by using the labeled set $(x_{i}^{l}, y_{i}^{l})$. Each prototype $\{h_{k}\}^{C}_{k=1}$ is a mean vector of the output of the encoder $g^{p}(.)$ for all input samples $x_{i}^{l}$ in the labeled set and class label $y_{i}^{l} = k$: 
\begin{equation}
    h_{k} = \dfrac{1}{N_{k}}\sum_{i=1}^{n_{l}}{g^{p}(x_{i}^{l})} 
\end{equation}

Every sample in the auxiliary dataset is then compared with each prototype to yield an affinity score $\{a_{i}\}^{n_{a}}_{i=1}$. We use cosine similarity as the distance measure between the two vectors. Other distance measures such a Euclidean distance could also be used in place of cosine similarity.
\begin{equation}
    a_{i} = \max_{k = [1, \dots, C]} \{sim(g^{p}(x_{i}^{a}), \; h_{k})\} 
\end{equation}
These scores show how similar auxiliary examples are to the labeled data prototypes, or their \textit{affinity} to the labeled data distribution.

\subsection{Consistency regularization and entropy maximization}
\label{sec:phase2}
A threshold $\tau$ on the affinity scores can be used to separate the auxiliary set into a selection set $D_{a}^{+}$ and regularization set $D_{a}^{-}$. We set the threshold according to the equation $\tau = \mu + \beta \sigma$ where $\mu$ and $\sigma$ are the mean and standard deviation of the affinity score distribution $\{a_{i}\}^{n_{a}}_{i=1}$ and $\beta$ is a hyperparameter:
\begin{equation}
\begin{split}
 D_{a}^{+} = \{x_{i}^{a}\} \; |\; a_{i} \geq \tau  
\\ D_{a}^{-} = \{x_{i}^{a}\} \; |\; a_{i} < \tau  
\end{split}
\end{equation}

Enforcing consistency in predictions across different augmentations of the same input image is a commonly used method in semi-supervised learning \cite{fixmatch-2020,mixmatch,remixmatch}. It encourages a label-transfer from the limited labeled data to the vast amount of in-distribution unlabeled data in general semi-supervised learning settings. We follow the same principle of consistency regularization for the labeled set $D_{l}$ and the selection set $D_{a}^{+}$. In particular, we implement the strategy used in methods such \cite{fixmatch-2020, uda, remixmatch}, which consider a weak and a strong augmentation of an input sample denoted by $\alpha(.)$ and $\mathcal{A}(.)$ respectively. The pre-trained encoder from the previous step $g^{p}(.)$ is now attached to a new $C$-class classifier head $c^{T}(.)$ to form our target model $f_{\theta}^{T}$. Weakly augmented samples from the labeled set are used to train $f_{\theta}^{T}$ with the standard supervised cross-entropy loss $\mathcal{L}_{x}$ over a batch $B$ of labeled samples as:
\begin{equation}
\mathcal{L}_{x} = \dfrac{1}{B} \sum_{i=1}^{B}H\big(y_{i}^{l}, f_{\theta}^{T}(\alpha(x_{i}^{l}))\big)
\end{equation}

Moreover, hard pseudo-labels are generated only for the weakly-augmented versions of samples for data from the selection set $D_{a}^{+}$ rather than all of the unlabeled set. The pseudo-labels $\hat{y}_{i}$ are used to enforce consistency on the strongly-augmented versions of the same input sample by minimizing the loss $\mathcal{L}_{u}^{+}$. Following \cite{fixmatch-2020} during training, for a batch $B$ of labeled data, $\mu B$ batches of unlabeled data are sampled and used.
\begin{equation}
\begin{split}
\hat{y}_{i} &= arg \max\big(f_{\theta}^{T}(\alpha(x_{i}^{a}))\big)  \\
\mathcal{L}_{u}^{+} &= \dfrac{1}{\mu B} \sum_{i=1}^{\mu B} H\big(\hat{y}_{i}, f_{\theta}^{T}(\mathcal{A}(x_{i}^{a}))\big) \quad \forall \; (x_{i}^{a}) \in D_{a}^{+}.
\end{split}
\end{equation}

Learning from the pseudo-labels of the selection set minimizes the entropy on the source label set $Y_{l}$. On the other hand, the data from the regularization set is used to regularize the model by maximizing the entropy of the output predictions. This min-max entropy optimization benefits the model by contrasting positive samples from the selection set against the negative samples from the regularization set.
Positive samples are relatively reliable, so they are used in the semi-supervised setting with consistency regularization. On the other hand, negative samples are not as reliable, so they are used to push the model to be unable to decide which class labels to assign to these samples. 
We penalize prediction confidences by assigning a uniform distribution over all classes as the target distribution for rejection samples. The loss $\mathcal{L}_{u}^{-}$ is then defined as:
\begin{equation}
\begin{split}
p(\hat{y}_{i} &= k) = \dfrac{1}{|Y_{L}|} \quad  \quad \forall \; k \in Y_{l} \\
\mathcal{L}_{u}^{-} &= \dfrac{1}{\mu B} \sum_{i=1}^{\mu B} \{ H\big(\hat{y}_{i}, f_{\theta}^{T}(\alpha(x_{i}^{a}))\big) +  H\big(\hat{y}_{i}, f_{\theta}^{T}(\mathcal{A}(x_{i}^{a}))\big)\} \\ & \quad  \quad \forall \; (x_{i}^{a}) \in D_{a}^{-}.
\end{split}
\end{equation}

Finally, the net loss minimized by the model is $\mathcal{L} = \mathcal{L}_{x} + \lambda_{u}^{+} \mathcal{L}_{u}^{+} + \lambda_{u}^{-} \mathcal{L}_{u}^{-}$, where $\lambda_{u}^{+}$ and $\lambda_{u}^{+}$ are loss weights for the selection set and the regularization set respectively. AuxMix maintains an Exponential Moving Average (EMA) \cite{meanteacher-2017} of the parameters $\theta$ of the model for stable training and inference in both phases.
\section{Experiments}
\label{sec:experiments}
In this section we review and discuss the experiment results, as well as ablation studies.

\subsection{Experiments settings}
We evaluate the performance of AuxMix on the standard semi supervised learning benchmark dataset, CIFAR10 \cite{cifar10}. 
The unlabeled data however, is drawn from entirely different datasets: (1) Tiny-ImageNet, (2) Caltech-UCSD Birds - 2011 (CUB) \cite{cub}, and (3) Noise images sampled from a uniform random distribution, to constitute our auxiliary set. We compare our results for each of these datasets with the recent SOTA SSL methods: FixMatch \cite{fixmatch-2020} and MPL \cite{MPL}. In addition, we report on several other SSL methods previously evaluated by \cite{realmix} and \cite{oliver} on the Animals-vs-Others class split of the CIFAR10 dataset. We also perform results of ablations to determine the contribution of each component of our algorithm. 

\subsection{Training details}
We use RandAugment \cite{randaug-2019} with parameters as defined in \cite{fixmatch-2020} as the strong augmentation and random horizontal flips and random crops as the weak augmentation for AuxMix. In all of our experiments, we set the following hyper-parameters: $\lambda_{u}^{+} = 1, \lambda_{u}^{-} = 1, B = 64, \mu=7$, ema-decay = 0.999, weight-decay=$5e-4$, rotation learning rate phase 1 = 0.1, learning rate phase 2 = 0.03, and cosine learning rate scheduling. All experiments including baselines are run for 300k iterations and we report the best score on 6k test set for CIFAR10-Animals-Others and 10k standard test set for all other experiments. The threshold $\tau$ is varied by changing the hyperparameter $\beta$ and the best results are reported in the tables.

\subsection{Distribution mismatch results on CIFAR10 Animals-vs-Others} 
We follow the distribution mismatch experiment settings as described in \cite{oliver} which splits the CIFAR10 dataset into 400 samples of the six animals classes as the labeled set and the rest of the data (20000 samples) from the four other classes as the auxiliary set. We report the results for our method and other baselines on Wide ResNet 28-2 \cite{wrn} in Table \ref{tab:cifar10-animals-others}. AuxMix at 14.12\% shows 3.5\% lower error rate than RealMix and over 8\% lower than supervised training with the labeled set only. It is interesting to note that most methods show a higher error rate than supervised learning due to the unlabeled data being out-of-distribution. That means unsuitable unlabeled data may in fact damage the learning.

\vspace{10pt}
\begin{table}
\vspace{10pt}
\begin{center}
\begin{tabular}{lc}
\toprule
Method                           & Error rate (\%) \\
\midrule 
Supervised                      & 22.47 \\
\hline
Temporal Ensembling \cite{TemporalEnsembling}     & 27.02 \\
Mean Teacher \cite{meanteacher-2017}            & 26.81 \\
VAT \cite{VAT}                                  & 26.19 \\
Pseudo-Label \cite{pseudolabel}                 & 25.94 \\
SWA \cite{SWA}                                  & 24.10 \\
UASD \cite{uasd}                                & 22.47 \\
RealMix \cite{realmix}                          & 17.62 \\
AuxMix (Ours)                                   & \textbf{14.12} \\
\bottomrule

\end{tabular}

\caption{\label{tab:cifar10-animals-others}Error rate on distribution mismatch between CIFAR10-Animals as labeled set and CIFAR10-Others as unlabeled set on Wide-ResNet-28-2. Lower is better.}
\end{center}
\end{table}

\subsection{Tiny-Imagenet, CUB and noise as auxiliary data}
Next, we investigate the impact of changing the 
auxiliary dataset. 
For the auxiliary set, we use 100k images across 200 classes from the training set of Tiny-Imagenet, 11788 images from CUB, and 50k noise samples drawn from a random uniform distribution, in three sets of experiments.
We use a 4K labeled random subset of CIFAR10 as the labeled set. Moreover, we use a ResNet-50 \cite{resnet} model to compare with previous work. All baselines are trained for 300k iterations except pseudo-labelling and supervised training, which we run for 100k iterations. Other hyper-parameters for each method are set according to the descriptions in their respective papers. 
Table \ref{tab:cifar10-resnet50} contains results from this experiment. We observe from this table that AuxMix performs competitively in comparison to other baselines, across 
different auxiliary datasets. AuxMix is close to 10\% better than FixMatch and close to 19\% better than MPL when TinyImagenet is used as the auxiliary dataset. It is interesting to note that pseudo-labelling performs very well when compared to the recent SSL baselines in most cases. In addition, as expected, using auxiliary data from outside label distributions in some baselines results in a performance lower than that of the supervised model.

\vspace{10pt}
\begin{table}
\vspace{10pt}
\begin{center}
\begin{tabular}{lccc}
\toprule
Method &  \multicolumn{3}{c}{Accuracy (\%)} \\
\cmidrule(l{3pt}r{3pt}){2-4}
{}  & Tiny-Imagenet & CUB       & Noise \\
\midrule
Supervised-4K & 59.91 & 59.91 & 59.91\\
Pseudo-Label \cite{pseudolabel}   &  63.04    &      61.80     &  64.89     \\
MPL~ \cite{MPL}                      &  49.18  &  68.86      &     52.44  \\
FixMatch~\cite{fixmatch-2020}       &  58.48    &      49.23     &     65.88  \\
AuxMix (Ours)                       &  \textbf{68.38}     &  \textbf{73.95}  &     \textbf{69.34}  \\
\bottomrule
\end{tabular}
\caption{\label{tab:cifar10-resnet50}
Classification accuracy for the 4K CIFAR10 experiment with the test set from CIFAR10 test set, and using entirely different datasets as unlabeled auxiliary data.} 
\end{center}
\end{table}


\subsection{Ablation study}
Table \ref{tab:ablation_study} shows an ablation study on AuxMix, where we study the contributions of the entropy-maximization and sample masking components of our method. For this study, we used the CIFAR10 Animals-vs-Others experiment settings.
As shown in Table \ref{tab:ablation_study}, each component on its own achieves a reasonable accuracy, but together the highest performance is achieved.

\vspace{10pt}
\begin{table}
\vspace{10pt}
\begin{center}
\begin{tabular}{lc}
\toprule
Method                           & Accuracy (\%) \\
\midrule 
Supervised                      &       77.53        \\
AuxMix                          &       85.88        \\
AuxMix without Entropy Maximization        &       85.10        \\
AuxMix without Sample Masking              &       84.57        \\
\bottomrule

\end{tabular}

\caption{\label{tab:ablation_study}
Ablation study on AuxMix with CIFAR10 Animals-vs-Others experiment. We observe that masking and regularization both contribute to the end-to-end performance.}
\end{center}
\end{table}

\section{Conclusion}
\label{sec:conclusion}

In this paper we first provided empirical evidence as well as intuitions that unlabeled data from unconstrained distributions can considerably damage the semi-supervised learning accuracy of existing SSL methods. Then we formalized this SSL setting as a new Auxiliary-SSL paradigm. Moreover, we proposed an algorithm called AuxMix to tackled the issues raised in Auxiliary-SSL. Experimental evaluations showed that AuxMix can achieve a competitive performance in recovering the performance drops occurred when using unconstrained unlabeled data.

\end{spacing}

{\small
\bibliographystyle{ieee_fullname}
\bibliography{references}
}

\end{document}